\documentclass[sigconf]{acmart}

\usepackage{booktabs} % For formal tables
\usepackage{caption}
\usepackage{amsmath}

\renewcommand\footnotetextcopyrightpermission[1]{} % removes footnote with conference information in first column
\begin{document}
\title{Highly Scalable Deep Learning Training System with Mixed-Precision: Training ImageNet in Four Minutes }

\author{\textbf{Xianyan Jia$^{*1}$, Shutao Song$^{*1}$, Wei He$^{1}$, Yangzihao Wang$^{1}$, Haidong Rong$^{1}$, Feihu Zhou$^{1}$,\\ Liqiang Xie$^{1}$, Zhenyu Guo$^{1}$, Yuanzhou Yang$^{1}$, Liwei Yu$^{1}$, Tiegang Chen$^{1}$, Guangxiao Hu$^{1}$,\\ Shaohuai Shi$^{*2}$, Xiaowen Chu$^{2}$}\\
Tencent Inc.$^{1}$, Hong Kong Baptist University$^{2}$\\
\{xianyanjia, sampsonsong, winsonwhe, slashwang, hudsonrong, hopezhou,\\ felixxie, alexguo, jokeyang, leolwyu, steelchen, gethinhu\}@tencent.com;\\ \{csshshi, chxw\}@comp.hkbu.edu.hk\\
*Authors contributed equally}

% \author{Xianyan Jia$^{*1}$, Shutao Song$^{*1}$, Wei He$^{1}$, Yangzihao Wang$^{1}$, Haidong Rong$^{1}$, Feihu Zhou$^{1}$, Liqiang Xie$^{1}$, Zhenyu Guo$^{1}$, Yuanzhou Yang$^{1}$, Liwei Yu$^{1}$, Tiegang Chen$^{1}$, Guangxiao Hu$^{1}$\\
% Tencent Inc.$^{1}$\\
% \{xianyanjia, sampsonsong, winsonwhe, slashwang, hudsonrong, hopezhou, felixxie,alexguo, jokeyang, leolwyu, steelchen, gethinhu\}@tencent.com}

% \author{Shaohuai Shi$^{2}$, Xiaowen Chu$^{2}$\\
% Hong Kong Baptist University$^{2}$\\
% \{csshshi, chxw\}@comp.hkbu.edu.hk\\
% *Authors contributed equally}

\begin{abstract}
%The ever-growing sizes of datasets and neural networks have put new challenges for high-performance training of deep neural networks. To achieve high performance and excellent scalability while maintaining small mini-batch accuracy, we have built an efficient high-performance distributed system for training neural networks. We use distributed synchronized stochastic gradient descent (SGD) to train mini-batches over a thousand GPUs with a total batch size of \textbf{64K}. In addition, we have adopted various optimization strategies such as mixed-precision training, layer-wise adaptive rates scaling (LARS), and customized collective communication operations with tensor fusion over Remote Direct Memory Access (RDMA) to further speedup the training procedure. Our TensorFlow-based system Jizhi can finish the \textbf{95-Epoch AlexNet training in 4 minutes with 58.7\% top-1 accuracy(batch size = 64K) using 512 Tesla P40 GPUs}, and finish the \textbf{90-Epoch ResNet-50 training in 7 minutes with 76.2\% top-1 accuracy(batch size = 64K) using 2048 Tesla P40 GPUs}. To the best of our knowledge, this is currently the \textbf{fastest} ImageNet training speed without losing accuracy.
Synchronized stochastic gradient descent (SGD) optimizers with data parallelism are widely used in training large-scale deep neural networks. Although using larger mini-batch sizes can improve the system scalability by reducing the communication-to-computation ratio, it may hurt the generalization ability of the models. To this end, we build a highly scalable deep learning training system for dense GPU clusters with three main contributions: (1) We propose a mixed-precision training method that significantly improves the training throughput of a single GPU without losing accuracy. (2) We propose an optimization approach for extremely large mini-batch size (up to 64k) that can train CNN models on the ImageNet dataset without losing accuracy. (3) We propose highly optimized all-reduce algorithms that achieve up to 3x and 11x speedup on AlexNet and ResNet-50 respectively than NCCL-based training on a cluster with 1024 Tesla P40 GPUs. On training ResNet-50 with 90 epochs, the state-of-the-art GPU-based system with 1024 Tesla P100 GPUs spent 15 minutes and achieved 74.9\% top-1 test accuracy, and another KNL-based system with 2048 Intel KNLs spent 20 minutes and achieved 75.4\% accuracy. Our training system can achieve 75.8\% top-1 test accuracy in only \textbf{6.6 minutes} using 2048 Tesla P40 GPUs. When training AlexNet with 95 epochs, our system can achieve 58.7\% top-1 test accuracy within \textbf{4 minutes}, which also outperforms all other existing systems.
\end{abstract}


\maketitle

\section{Introduction}
With the ever-increasing sizes of datasets and larger deep neural networks, training often takes several days if not weeks (For example, training ResNet-50~\cite{he2016deep} takes 29 hours using 8 Tesla P100 GPUs). Extremely long training time impedes the research and development progress. Due to the single machine's limited computing resources, it is natural to distribute the workload to clusters and use supercomputing power to increase the throughput of data flow. A commonly adopted solution is distributed synchronous Stochastic Gradient Descent (SGD) which parallelizes the tasks across machines. To make full use of the hardware, mini-batch size per machine should be properly set and cannot be too small. In addition, it is common to use large batch to achieve weak scaling ~\cite{goyal2017accurate,cho2017powerai,you2017imagenet,akiba2017extremely,codreanu2017scale,smith2017don}. In this way, the speedup is obtained by utilizing overall throughput of the system and fewer updates of the model. 

However, there are two challenges when using large batch across large clusters:
\begin{itemize}
   \item \textbf{Challenge 1}: Larger mini-batch size often leads to lower test accuracy, as there exists a generalization gap~\cite{keskar2016large}.
    \item \textbf{Challenge 2}: When using large clusters, it is harder to achieve near-linear scalability as the number of machines increases, especially for models with the high communication-to-computation ratio.
\end{itemize}

\textbf{Challenge 1}. Larger mini-batch size reduces the variance of gradients by taking the average of the gradients in mini-batch, and it provides a more accurate estimate of the gradients~\cite{goodfellow2016deep}. Thus it allows the model to take bigger step size and in turn makes the optimization algorithm progress faster. However, as reported in~\cite{you2017imagenet}, when increasing the mini-batch size to 64K, the test accuracy of ResNet-50 drops from 75.4\% to 73.2\%. \citet{codreanu2017scale} also train ResNet-50 with batch 64K and achieves the accuracy of 73.9\%, which does not meet the baseline accuracy.

\textbf{Challenge 2}. A distributed training system with data parallelism strategy typically divides batches across each GPU, and requires a gradient aggregation step between each training step. This communication step usually becomes the bottleneck of the system when the number of GPUs becomes large. To achieve high performance for such distributed training system, we need to improve both the single GPU performance and the overall system performance. Given that the training throughput with one GPU is $S$, if we use $N$ GPUs with the scaling efficiency $e$, then the system throughout should be $T=S \cdot N \cdot e$. When the number of GPUs $N$ is fixed, we need to increase both $S$ and $e$ to improve the overall throughput of the training system $T$. To improve the throughput, we need faster computation and more efficient bandwidth utilization, to improve the scaling efficiency, we need more efficient collective communication primitives that can handle a system with thousands of GPUs.

In this paper, we have addressed the above two challenges. Our contributions are as follows:
\begin{itemize}
    \item We successfully scale the mini-batch size to 64K for AlexNet and ResNet-50 training without loss of accuracy. To achieve this, we have adopted and proposed several strategies (i.e., mixed-precision training with LARS, eliminated weight decay on bias and parameters for batch normalization, and adding proper batch normalization layers).
    \item We build a high-throughput distributed deep learning training system which contains two main features to improve the single GPU performance $S$ and the system scaling efficiency $e$. 1) To improve $S$, our system supports half-precision training, which theoretically, could achieve two times throughput improvement compared to its single-precision counterpart. 2) To improve $e$, our system uses a hybrid strategy which combines our optimized adaptive all-reduce collective with the ring-based all-reduce in NCCL\@.
\end{itemize}

The rest of this paper is structured as follows. We show the related work in Section~\ref{sec:related_work}, then describe the design and implementation of our system and optimizations in Section~\ref{sec:sys_overview} and Section~\ref{sec:sys_impl}. Finally, we discuss our experimental results in Section~\ref{sec:exp} and conclude with experiences we have learned through building such a system in Section~\ref{sec:conc}.

\section{Related Work}
\label{sec:related_work}
This section describes the research landscape of distributed deep learning training system in three fields: 1) large-batch training; 2) low-precision training; 3) distributed training on heterogeneous clusters.

\subsection{Large-batch Training}
\citet{goyal2017accurate} first trains the ResNet-50 ImageNet model with a large mini-batch size of 8K over 256 Tesla GPUs and finishes the training process within one hour. They adopt the linear scaling rule to adjust the learning rate as a function of mini-batch size. They also develop a warmup scheme, i.e., starting from small learning rate and slowly increasing the learning rate through a few epochs, in order to overcome the optimization challenges in the first few epochs. \citet{cho2017powerai} use the same training configuration and finish ResNet-50 training in 50 minutes with 256 GPUs. \citet{you2017imagenet} further increase the mini-batch size of ResNet-50 from 8K to 32K\@. They use LARS to enable large mini-batch size and can finish the ResNet-50 training in 20 minutes with 2048 KNL chips. Besides ResNet-50, they also experiment on AlexNet and finish the training of mini-batch size 32K on ImageNet in 11 minutes with 1024 Skylake CPUs. \citet{akiba2017extremely} demonstrate the training of ResNet-50 in 15 minutes with a mini-batch size of 32K over 1024 Tesla P100 GPUs. They adopt techniques such as RMSprop warm-up, batch normalization without moving average and a slow-start learning rate schedule. However, their reported test accuracy is 74.9\%,  which is lower than the baseline $\sim$75.3\%.  \citet{codreanu2017scale} use a combination of techniques such as aggressive learning rate scheduling and improved weight decay. They reach 75.3\% test accuracy using 32K mini-batch size in 42 minutes and 74.6\% test accuracy using 49K mini-batch size in 28 minutes. In addition to adjusting the learning rate, \citet{smith2017don} propose to increase the mini-batch size instead of decaying the learning rate. Their work is the first to train ImageNet with less time (30 minutes) without losing accuracy after \citet{goyal2017accurate}. All the above research towards large-batch training either fails to scale to more nodes and more GPUs with larger mini-batch size, or trade accuracy loss for better performance.

\citet{devarakonda2017adabatch} use dynamic mini-batch size and decay the learning rate at the same time. However, adapting the mini-batch size is only tested in piece-wise constant learning rate schedule, but cannot be easily applied in polynomial decay, whose curve is more smooth.

\subsection{Low-precision Training}
Low-precision computation is often used to lower the time and energy cost of machine learning. Unfortunately, the benefits of low-precision (LP) arithmetic come with a cost. The round-off or quantization error that results from converting numbers into a low-precision representation introduces noise that can affect the convergence rate and accuracy of SGD\@. Conventional wisdom says that, for training, low-precision introduces a tradeoff of the number-of-bits used versus the statistical accuracy: the fewer bits used, the worse the solution will become. Theoretical upper bounds on the performance of low-precision SGD [9] and empirical observations of implemented low-precision algorithms~\cite{courbariaux2015multi} further confirm that current algorithms are limited by this precision-accuracy tradeoff. \citet{christopherde2018high} describe a simple low-precision stochastic gradient descent variant called HALP, which converges at the same theoretical rate as full-precision algorithms despite the noise introduced by using low precision throughout execution. The key
idea is to use Stochastic Variance Reduced Gradient (SVRG) to reduce gradient variance, and to combine this with a novel technique called bit centering to
reduce quantization error. \citet{micikevicius2017mixed} propose three techniques for preventing the loss of critical information. Firstly,
they recommend maintaining a single-precision copy of weights that accumulates the gradients after each optimizer step (this copy is rounded to half-precision for the forward- and back-propagation). Secondly, they propose loss-scaling to preserve gradient values with small magnitudes. Although the loss-scaling technique was not a requirement for successful mixed-precision training when mini-batch size is not large enough. Thirdly, they use half-precision arithmetic that accumulates into single-precision outputs, which are converted to half-precision before storing to memory. While all tensors in the forward and backward passes were in FP16 format, a master copy of weights was updated in FP32 format. However, they have not applied this with large-batch training strategy such as LARS to achieve better performance.

\subsection{Distributed Training on Heterogeneous Clusters}
Most distributed machine learning frameworks such as TensorFlow~\cite{marting2015tensorflow} adopt centralized deployment mode. One bottleneck of the centralized algorithm is the high communication cost on the central nodes. Baidu~\cite{baidu2017ringallreduce} first introduced the ring-based all-reduce algorithm~\cite{barnett1994ring} to deep learning. This is a very important contribution to the field of distributed training. The ring all-reduce algorithm greatly reduces the communication load when the number of nodes increases. However, the original version is low in bandwidth utilization because of splitting up the tensors data into too small slices when tensor sizes are small compared to the number of nodes in the cluster. The IBM's PowerAI Distributed Deep Learning (DDL) system~\cite{cho2017powerai} has mentioned a new all-reduce algorithm. However, since the implementation is closed source, it is difficult to be applied to other works. \citet{goyal2017accurate} use an implementation of all-reduce consists of three phases for intra-node and inter-node communication: 1) intra-node reduce, 2) inter-node all-reduce, and 3) intra-node broadcast, which reduces the communication load across nodes and improves scalability. Horovod~\cite{sergeev2018horovod} introduced gradient fusion strategy to the all-reduce algorithm, which reduces tensor fragmentation and improves bandwidth utilization. However, this indiscriminate fusion results in unnecessary memory copy and no significant gains have been made in our test scenario. The DAG model proposed by \citet{shi2018modeling} for scheduling the computation and communication tasks in synchronized SGD guides us to design our optimized all-reduce algorithm.

Our system adopts several useful parts from the above work. Together with other optimizations, they help us yield high scalability on ImageNet training with both AlexNet and ResNet-50.

\section{System Overview}
\label{sec:sys_overview}
\begin{figure}
    \centering
    \includegraphics[width=1\linewidth]{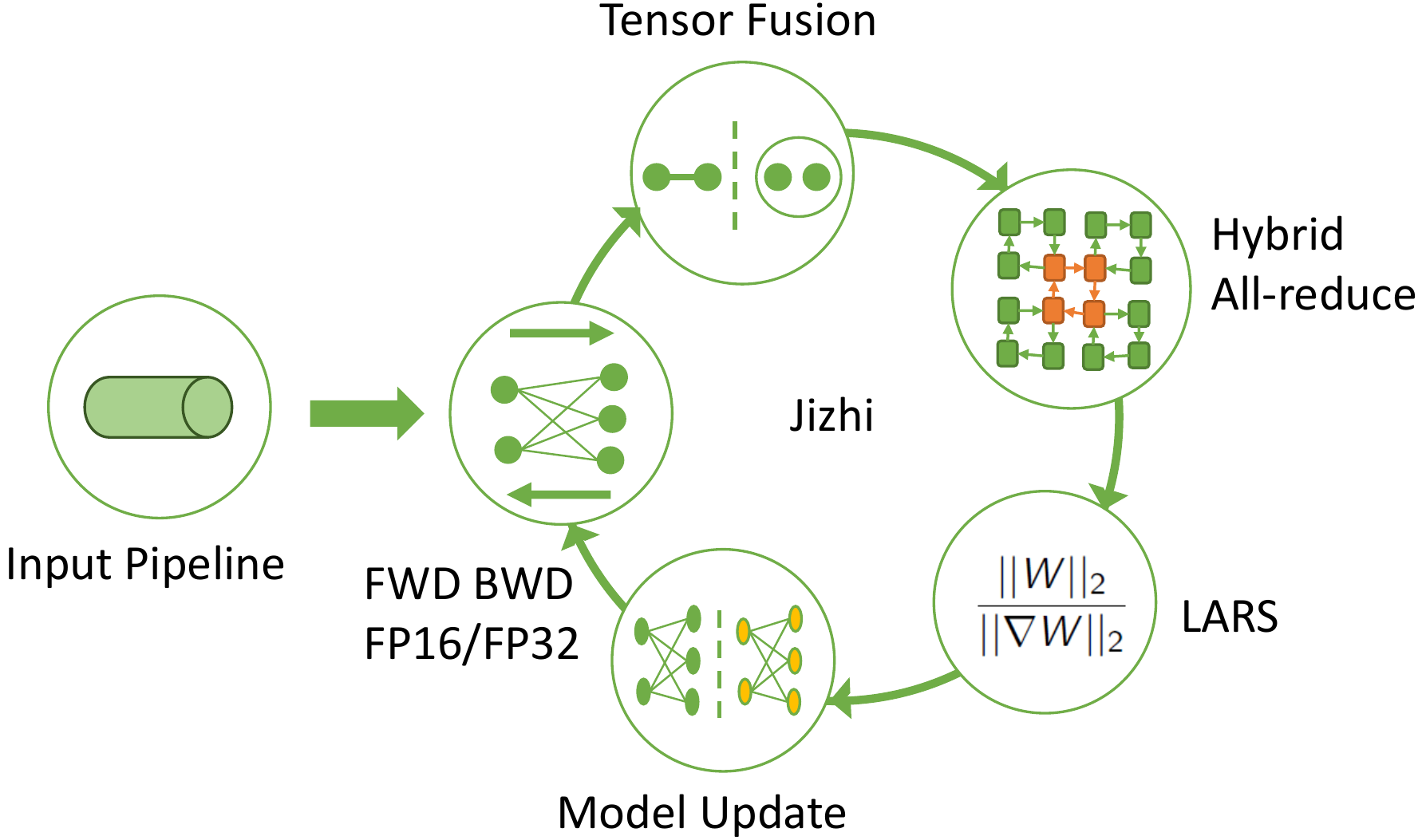}
    \caption{Jizhi Training System Overview}
    \label{fig:system}
\end{figure}

Figure~\ref{fig:system} is an overview of our distributed deep learning training system. At a high level, our system contains the following three modules: 1) input pipeline module; 2) training module; and 3) communication module. 
\begin{itemize}
    \item  The input pipeline module delivers data for the next step before the current step has finished. It uses pipelining in order to minimize both CPU and GPU idle time.
    \item The training module includes model construction and variable management. In this module, we have incorporated optimizations such as forward/backward computation with mixed-precision and model update with LARS\@.
    \item The communication module uses tensor fusion and hybrid all-reduce to optimize the scaling efficiency according to tensor size and cluster size.
\end{itemize}

\section{System Implementation and Optimizations}
\label{sec:sys_impl}
\subsection{Mixed-Precision Training with LARS}
\label{sec:Mixed-Precision}
As \citet{micikevicius2017mixed} have mentioned, the motivation of using half-precision (FP16) in the training phase is to lower memory bandwidth pressure as well as increase arithmetic throughput. The former can be achieved by using fewer bits to store the same number of values, the latter is achieved on processors that offer higher throughput for reduced precision math.
Orthogonal to half-precision training, \citet{you2017Scaling} first proposed LARS to enable larger mini-batch size for distributed training. The algorithm introduces a local learning rate for each layer (as shown in Equation~\ref{eq:lars}), which is the ratio of the L2-norm of weights and gradients weighted by a LARS coefficient $\eta$. Gradients are multiplied with its adaptive local learning rate.
A natural choice is to combine half-precision training with LARS to achieve larger mini-batch size with scalability. However, a na{\"i}ve implementation would introduce several problems because using LARS directly on half-precision training will cause the computed learning rate to be out of the dynamic range of IEEE half-precision format (FP16), and thus cause the gradients to vanish and stall the training process.

    \begin{equation}
        \Delta w_{t}^{l} = \gamma \cdot \eta \cdot \frac{\lVert w^l \rVert}{\lVert \bigtriangledown L(w^l))\rVert} \cdot \bigtriangledown L( w_{t}^{l})
        \label{eq:lars}
    \end{equation}
    
\begin{figure}[ht]
    \centering
    \includegraphics[width=1\linewidth]{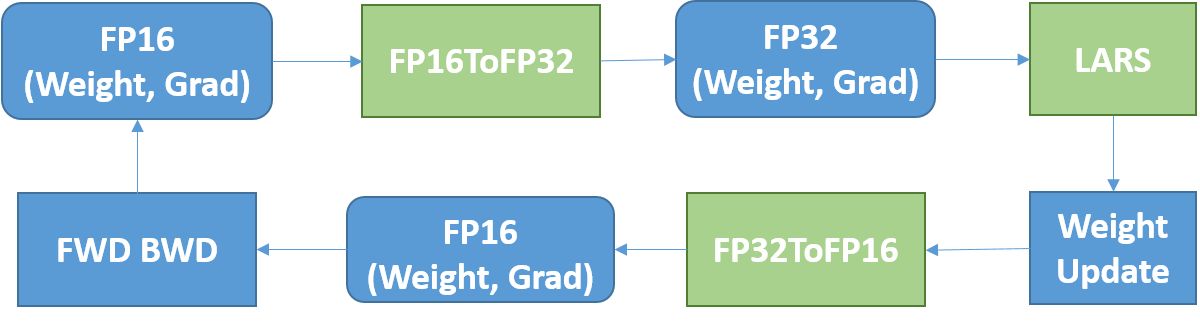}
    \caption{Mix-precision training with LARS}
    \label{fig:mix}
\end{figure}

To cope with this situation, we have proposed a training strategy which uses mixed-precision training with LARS as shown in Figure~\ref{fig:mix}. In our strategy, the operations in forward and backward propagation are performed in FP16, while the weights and gradients are cast to single-precision (FP32) format before applying LARS and cast back to FP16 afterward.

Mixed-precision training with LARS is one of the critical reasons that our system could keep good scalability while increasing the mini-batch size to 64K\@. Table~\ref{tb-lars} shows that on ResNet-50 with the mini-batch size of 64K, using LARS with mixed-precision training could maintain the top-1 accuracy as 76.2\%.

\begin{table}[ht]
  \caption{Effectiveness of using LARS on ResNet-50}
  \label{tb-lars}
  \centering
  \begin{tabular}{ccccc}
    \toprule
    Mini-Batch Size     & Number of Epochs  & LARS   & Top-1 Accuracy \\
    \midrule
    64K   & 90 & NO  & 73.2\%   \\
    64K   & 90 & YES   & 76.2\%  \\
    \bottomrule
  \end{tabular}
\end{table}

\subsection{Improvements on Model Architecture}
Improvements in model architecture could often lead to better performance. In our system, we have improved the model architecture from the following two aspects: 1) eliminated weight decay on the bias and batch normalization; and 2) adding proper batch normalization layers for AlexNet.

Weight decay is a commonly-used strategy to get better generalization for the model by adding a regularization term to the loss function $E$~\cite{krogh1992weightdecay}.
\begin{equation}
        E(w) = E_0(w) + \frac{1}{2}\lambda\sum_i{w_i^2}
        \label{eq:l2}
\end{equation}
%insert weight decay equation and describe its importance for generalization.
If gradient descent is used for learning, the last term of the loss function leads to a new term $-\lambda w_i$ in the gradients update:
\begin{equation}
    w_i^{t+1} = w_i^{t} - \eta \frac{\partial E}{\partial w_i^{t}} - \lambda w_i^{t}
\end{equation}

%say why we leave out bias and BN alpha/beta and it increases the performance.
In neural network training, it is a typical practice to penalize only the weights of the affine transformation at each layer and leaves the biases unregularized~\cite{goodfellow2016deep}. What we have observed in our training for AlexNet is that if we also leave two parameter sets $\beta$ and $\gamma$ in batch normalization unregularized, our model could achieve better convergence and usually with less time to train for the same number of epochs. $\beta$ and $\gamma$ are two trainable parameters in batch normalization as shown in below formulas, where $\mu_\mathcal{B}$ is the mean of mini-batch and $\sigma^2$ is the variance of mini-batch. $\beta$ and $\gamma$ contrals the scale and shift of the normalized result.
      $$ \hat{x}_i \leftarrow \frac{x_i-\mu_\mathcal{B}}{\sqrt{\sigma^2 + \epsilon}}$$
      $$  y_i \leftarrow \gamma \Hat{x}_i + \beta \equiv BN_{\gamma, \beta}(x_i) $$

As \citet{goodfellow2016deep} has noted, the reason that our model achieves better convergence could be that $b$, $\beta$ and $\gamma$ usually have much less parameters compared to weights $W$ (for AlexNet model, $b$, $\beta$ and $\gamma$ parameters only amount to 0.02\% of all parameters), which means that leaving them unregularized would not give too much variance and regularizing could instead, introduce a significant amount of underfitting. As shown in Table~\ref{tb-alexnet-decay}, for AlexNet, we get around 1.3\% improvement in accuracy with the same number of epochs. The slight improvements of training run time come from the reduced computations in L2 regularization.
%Instead of apply weight decay to all trainable variables in ResNet-50 and AlexNet (weights, biases, batch normalization (BN) beta $\beta$ and BN gamma $\gamma$), we only penalize the weights and leave the others unregularized. As shown in Equation~\ref{eq:bias}, to fit the model, $W_{ij}$ needs to interact with two variables ($y_i$, $x_j$), while $b_i$ only needs to interact with $y_i$. Since the pair ($y_i$, $x_j$) has more possible combinations than the single value $y_i$, we need more data to fit $W$ correctly than $b$. On the other hand, $b$, $\beta$ and $\gamma$ usually have much less parameters compared to weights $W$(for AlexNet model, $b$, $\beta$ and $\gamma$ parameters only amount to 0.02\% of all parameters), they are not the reason for overfitting. Not restricting $b$, $\beta$ and $\gamma$ will make the network more flexible and reduce some computations. As shown in Table~\ref{tb-alexnet-decay}, we get around 1.3\% improvement by excluding $b$, $\beta$ and $\gamma$ from weight decay.
% 2) Regularization restricts the parameters to grow large in magnitude. Allowing biases to grow large in magnitude may allow neurons to saturate faster.
%\begin{equation}
%    y_i = x_1W_{i1}+x_2W_{i2}+...+x_nW_{in} + b_i
%    \label{eq:bias}
%\end{equation}

\begin{table}[ht]
  \caption{Effect of Regularization with $b$, $\beta$ and $\gamma$ for AlexNet}
  \label{tb-alexnet-decay}
  \centering
  \begin{tabular}{ccccc}
    \toprule
    Batch     & Epochs  & Regularize $b$, $\beta$ and $\gamma$   & Top1 \\
    \midrule
    64K   & 95 &  Yes & 55.8\%   \\
    64K   & 95 &  No & 57.1\%  \\
    \bottomrule
  \end{tabular}
\end{table}

\begin{figure}[htbp]
    \centering
    \includegraphics[width=0.4\textwidth]{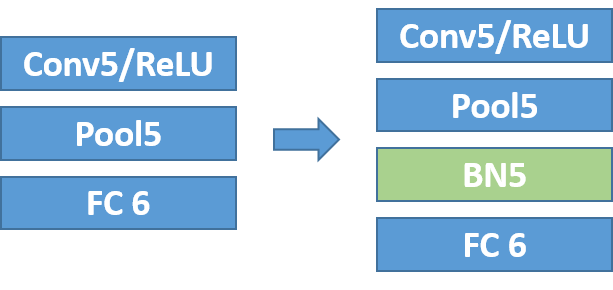}
    \caption{Insert Batch Norm after Pool5 for AlexNet}
    \label{fig:alexnet-bn}
\end{figure}

\begin{figure}[ht]
    \centering
    % \captionsetup{justification=centering}
    \includegraphics[width=0.5\textwidth]{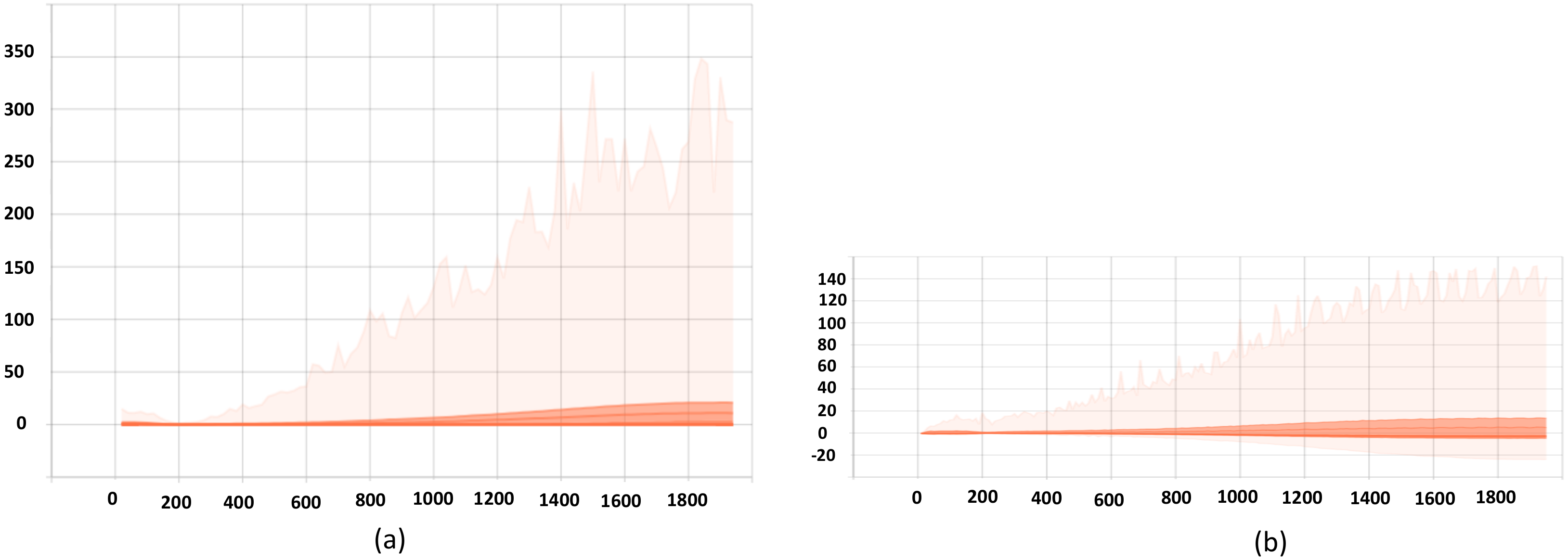}
    \caption{Feature Map Distribution of Pool5(a) and Pool5-BN5(b) of AlexNet as shown in Figure~\ref{fig:alexnet-bn}. (the horizontal axis is the training steps, the vertical axis is the feature map distributions.)}
    \label{fig:pool5-output}
\end{figure}

As mentioned in LARS~\cite{you2017Scaling}, replacing Local Response Normalization layers with Batch Normalization(BN) could improve the accuracy of AlexNet~\cite{ioffe2015batch}. However, as shown in Table~\ref{tb-alexnet-decay}, such AlexNet-BN model cannot reach baseline top-1 accuracy when the mini-batch size increases to 64K\@. By analyzing the parameters and feature map distributions, we find that the feature map distribution after Pool5 has a larger variance and maximum values as training go on (as shown in Figure~\ref{fig:pool5-output}(a)). The significant change of feature scaling makes the training difficult. This motivates us to insert another BN layer after Pool5 to rescale the feature map as shown in Figure~\ref{fig:alexnet-bn}. The refined-AlexNet-BN model could reach 58.8\% top-1 accuracy with 64K mini-batch size for 95 Epoch training.

\subsection{Improvements on Communication Strategies}
\begin{figure}[htbp]
    \centering
    \captionsetup{justification=centering}
    \includegraphics[width=0.5\textwidth]{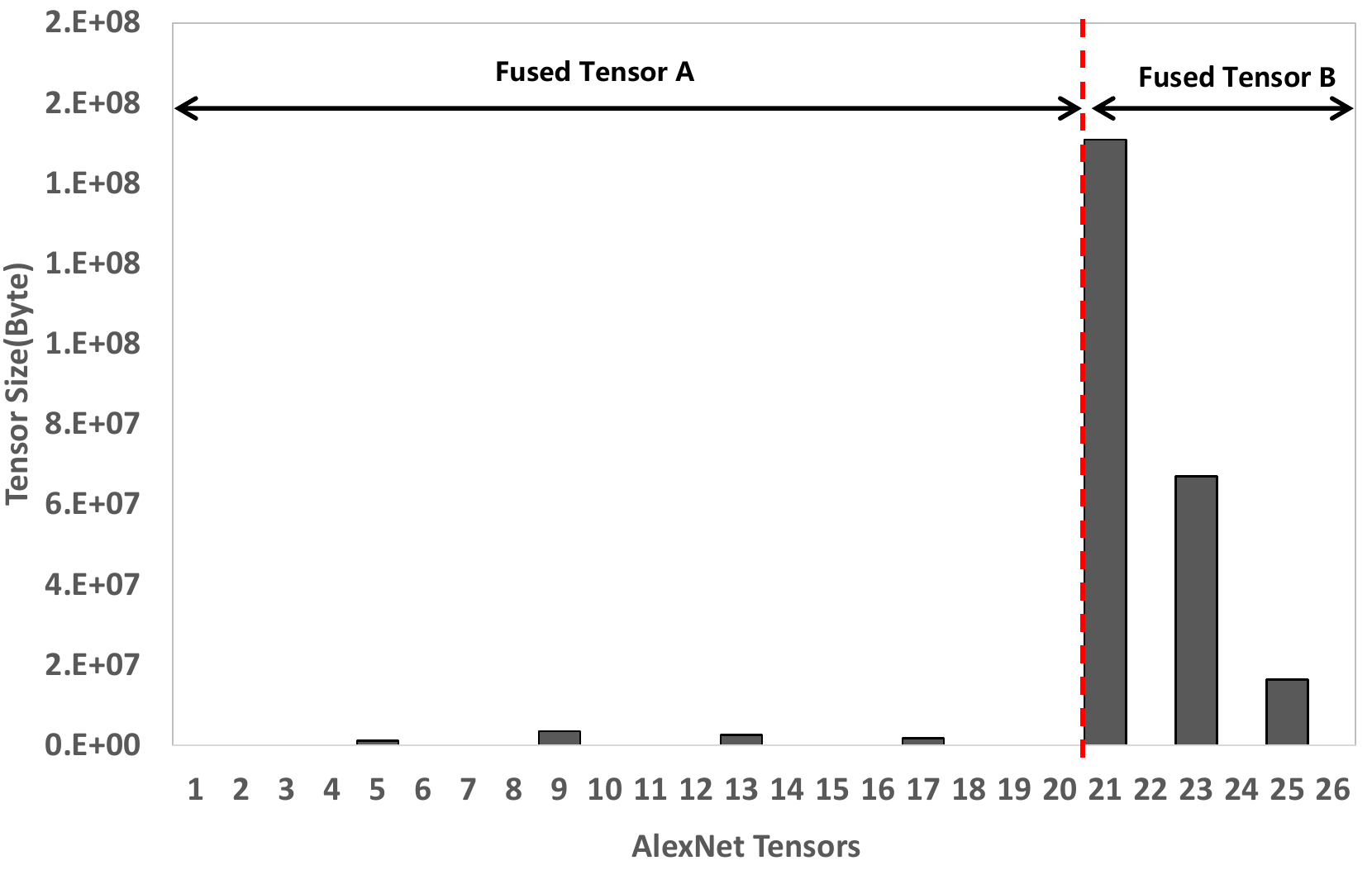}
    \caption{Tensor Fusion of AlexNet}
    \label{fig:fusion-alexnet}
\end{figure}

\begin{figure}[htbp]
    \centering
    \captionsetup{justification=centering}
    \includegraphics[width=0.5\textwidth]{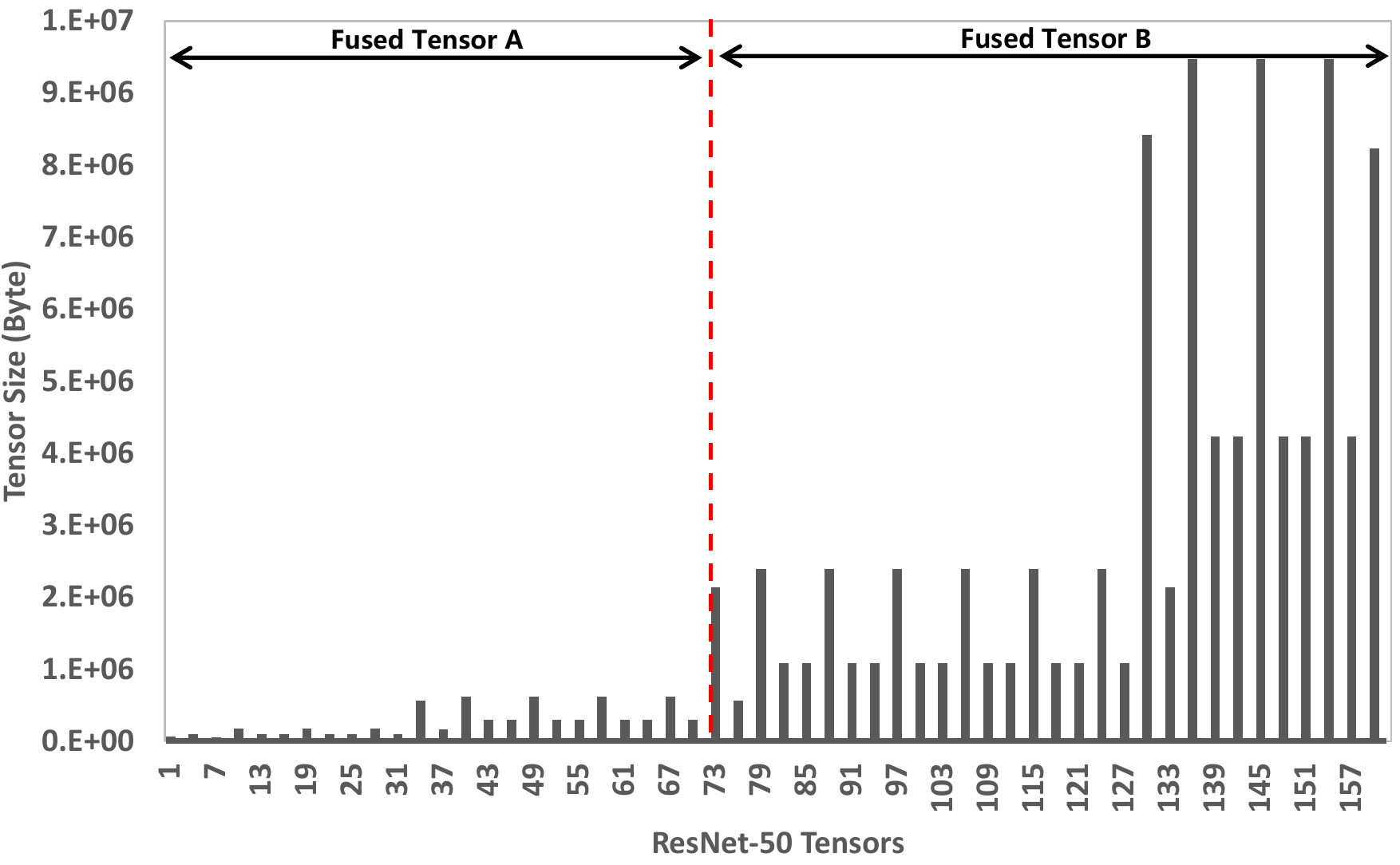}
    \caption{Tensor Fusion of ResNet-50}
    \label{fig:fusion-resnet}
\end{figure}

%comm is important for large-batch distributed sync SGD
For large batch training with distributed synchronized SGD, efficient gradients aggregation across all GPUs after each iteration is crucial to the training performance \cite{watcharapichat2016ako}\cite{shi2017performance}. \citet{goyal2017accurate} have pointed out that for models with a larger set of parameters and GPUs with more computing power, it becomes harder to hide the cost of aggregation in the backprop phase. In our case, for the mini-batch size of 64K and 1024 GPUs, gradients aggregation using collective communication primitives such as all-reduce has become the bottleneck of the system.
%problems for nccl and other implementations for large batch and multi-node
NCCL 2.0 is optimized for dense multi-GPU systems such as NVIDIA's DGX-1. In our case, communication happens over a hundred of nodes on a cluster, the traditional ring-based all-reduce implementation does not scale due to the following reason: In a cluster with $k$ GPUs, Ring all-reduce will split the data on each GPU into $k$ chunks and do the reduce in $k-1$ iterations \cite{thakur2005optimization}. When $k$ gets larger, the messages passing between nodes will become smaller and fail to utilize the full bandwidth of the network.
%our implementation
To cope with this problem, we have developed two strategies:
\begin{itemize}
\item \emph{Tensor Fusion.}
An efficient communication strategy in a distributed training system should maximize the throughput as well as reduce the latency. The main challenge of training deep neural networks with multiple layers is that the sizes of gradient tensors to aggregate vary a lot for different types of layers. Usually, gradient tensor sizes for convolution layers are much smaller than fully-connected layers. Sending too many small tensors in the network will not only cause the bandwidth to be under-utilized but also increase the latency. To cope with this problem, we adopt the technique of tensor fusion. The core idea of tensor fusion is to pack multiple small size tensors together before all-reduce to better utilize the bandwidth of the network. We set a parameter $\theta$. In the backward phase, as tensors from each layer come in, we fuse them into a buffer pool if the total size is less than $\theta$, and only send the fused tensor out for all-reduce when the total size is larger than $\theta$. This strategy could be easily generalized to distributed training for other neural networks. Figure~\ref{fig:fusion-alexnet} and Figure~\ref{fig:fusion-resnet} shows the fusion strategy for AlexNet and ResNet-50 respectively.

\item \emph{Hierarchical All-reduce.}
In our experiments for ResNet-50, when using tensor fusion to combine all the tensors into a single tensor, the end-to-end performance will increase by 8x. However, the high throughput also increases the latency, since fusing into a single tensor will prevent the parallelization of gradient aggregation of last few layers and backward propagation of earlier layers. To reduce latency, we need to restrict the condition for tensor fusion, i.e. allow smaller, and multiple tensors in tensor fusion phase. However, ring all-reduce perform worse on small tensors. Hierarchical all-reduce could solve this problem for small tensor communication. Instead of using ring all-reduce where each GPU sends and receives $\frac{m}{p}$ bytes of data in $2(p-1)$ steps. We can group $k$ GPUs together, then use a three-phase algorithm to do the all-reduce across all GPUs (Figure~\ref{fig:ring}: first we do a reduce within GPUs of the same group, store the partial results to a master GPU in each group, then we launch Ring all-reduce across $p/k$ groups, after each master GPU gets the final result, we do a broadcast within each group to propagate the final result to every GPU\@. The three-phase algorithm reduces the running steps from $2(p-1)$ to $4(k-1) + 2(p/k-1)$ since the intra-group reduce and broadcast each costs $2(k-1)$ steps. The decrease of computation steps makes the three-phase algorithm perform better in latency-sensitive case (i.e. for small tensor size and the large number of GPUs). We set $k$ as a tunable parameter and observe the highest performance is achieved when $k$ is set to 16 in our 1024 GPU cluster.
\begin{figure}[htbp]
    \centering
    \captionsetup{justification=centering}
    \includegraphics[width=0.5\textwidth]{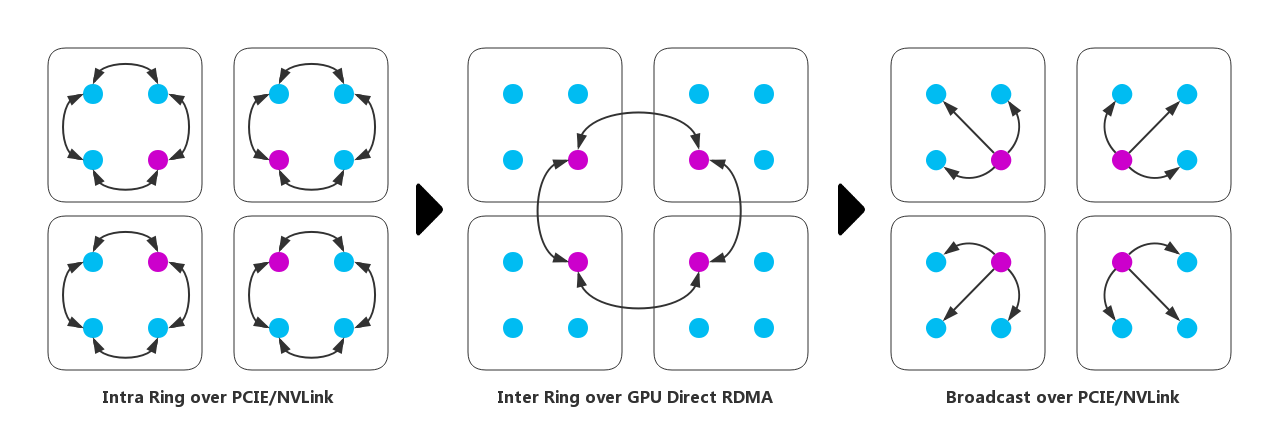}
    \caption{Our three-phase all-reduce algorithm for multi-node multi-GPU gradient aggregation.}
    \label{fig:ring}
\end{figure}
\item \emph{Hybrid All-reduce.}
Hierarchical all-reduce can bring performance gain for convolution layers which usually have a smaller number of weights. However, for fully-connected layers which usually have a much larger number of weights, ring-based all-reduce still outperforms our hierarchical all-reduce. To enjoy the best of both worlds, we use a hybrid strategy in our system. We set a parameter $\eta$ to represent the size of the tensor to aggregate in bytes. By tuning this parameter, we can switch between the traditional ring-based all-reduce and our customized all-reduce. Combined with tensor fusion, hybrid all-reduce could help us achieve better performance.
\end{itemize}

\section{Experimental Results}
\label{sec:exp}

\subsection{Experimental Settings}

\textbf{Model}.
We choose AlexNet~\cite{krizhevsky2012imagenet} and ResNet-50~\cite{he2016deep} for our experiments because they represent two typical types of CNN: As shown in Table~\ref{tb-model}, the parameter size of AlexNet is around 2.5 times as ResNet-50, while the computation of ResNet-50 is around 5.6 times as AlexNet. Thus, the bottleneck of AlexNet lies in communication, while the bottleneck of ResNet-50 lies in computation.
The baseline top-1 accuracy of AlexNet is 58.8\%~\cite{you2017imagenet} and the baseline top-1 accuracy of ResNet-50 is 75.3\%~\cite{he2016deep}.
% The above two models shows two typical aspects of distributed training difficulties.

\begin{table}[ht]
  \caption{Model Information}
  \label{tb-model}
  \centering
  \begin{tabular}{ccccc}
    \toprule
    Model     & Input Size     & Parameter Size & FLOPs & Baseline Top1\\
    \midrule
    AlexNet     & 227x227 &  62M    & 727 M  & 58.8\% \\
    ResNet-50     & 224x224 &  25M    & 4 G  & 75.3\% \\
    \bottomrule
  \end{tabular}
\end{table}

\textbf{Dataset}.
We use ImageNet~\cite{deng2009imagenet} dataset in the following experiments. Both models are trained on 1.28 million training images and evaluated on 50,000 validation images by top-1 test accuracy for its 1000 classes task. The training images are partitioned into 1024 chunks and the validation images are partitioned into 128 chunks. Images are stored in the format of TFRecord. In all our experiments, we use data augmentation offered in TensorFlow.

\textbf{Software}.
We use TensorFlow~\cite{abadi2016tensorflow} as a training framework for its flexible design, various use cases, and a large user/developer community. We build our distributed gradient aggregation algorithm with NVIDIA Collective Communication Library (NCCL), and OpenMPI\@.

\textbf{Hardware}.
Our GPU cluster includes 256 nodes, and each node contains 8 NVIDIA Tesla P40 GPUs that are interconnected with PCIe. For local storage, each server has two 2T NVMe SSDs.  For network connectivity, each server has a Mellanox ConnectX-4 100Gbit Ethernet network card. 
We use RoCEv2 (RDMA over Converged Ethernet) for communications among nodes in cluster, which is a common Remote Direct Memory Access (RDMA) implementations \footnote{RDMA is a technology which supports zero-copy networking by enabling the network adapter to transfer data directly to or from application memory, eliminating the need to copy data between application memory and the data buffers in the operating system.}. We also use GPUDirect RDMA (GDR) to enable direct data exchange between GPUs on different nodes. All of these technologies can reduce the latency and increase the scaling efficiency in our cluster.
\subsection{Overall experimental results}

\begin{table*}
  \caption{Compare AlexNet training with different teams}
  \label{tb-alexnet}
  \centering
  \begin{tabular}{c|ccc|cc}
    \toprule
    Team     & Batch  & Hardware & Software & Top-1 Accuracy & Time \\
    \midrule
    \citet{you2017imagenet}  & 512  & DGX-1 station & NVCaffe & 58.8\%    &  6h 10m \\
    \citet{you2017imagenet}  & 32K  & CPU $\times$ 1024 & Intel Caffe & 58.6\%    &  11min \\
    This work     & \textbf{64K} & Tesla P40 $\times$ 512 & TensorFlow      & \textbf{58.8\%} & \textbf{5m}\\
    This work     & \textbf{64K} & Tesla P40 $\times$ 1024 & TensorFlow      & \textbf{58.7\%} & \textbf{4m}\\
    \bottomrule
  \end{tabular}
\end{table*}

\begin{table*}
  \caption{Compare ResNet-50 training with different teams}
  \label{tb-resnet}
  \centering
  \begin{tabular}{c|ccc|cc}
    \toprule
    Team     & Batch  & Hardware & Software & Top-1 Accuracy & Time \\
    \midrule
    \citet{he2016deep} & 256 & Tesla P100 $\times$ 8 & Caffe & 75.3\% & 29h \\
    \citet{goyal2017accurate} & 8K & Tesla P100 $\times$ 256 & Caffe2 & 76.3\% & 1h\\
    \citet{cho2017powerai} & 8K & Tesla P100 $\times$ 256 & Torch & 75.0\% & 50min \\
    \citet{codreanu2017scale} & 32K & KNL $\times$ 1024 & Intel Caffe & 75.3\% & 42min \\
    \citet{you2017imagenet}  & 32K  & KNL $\times$ 2048 & Intel Caffe & 75.4\%    &  20min \\
    \citet{akiba2017extremely} & 32K & Tesla P100 $\times$ 1024 & Chainer & 74.9\% & 15min \\
    This work     & \textbf{64K} & Tesla P40 $\times$ 1024 & TensorFlow      & \textbf{76.2\%} & \textbf{8.7m}\\
    This work     & \textbf{64K} & Tesla P40 $\times$ 2048 & TensorFlow      & \textbf{75.8\%} & \textbf{6.6m}\\
    \bottomrule
  \end{tabular}
\end{table*}

For ResNet-50, as shown in Table~\ref{tb-resnet}, our system finishes the training in only 8.7 minutes with 76.2\% top-1 accuracy over 1024 Tesla P40 GPUs and 6.6 minutes with 75.8\% top-1 accuracy over 2048 Tesla P40 GPUs, which to our best knowledge is the state-of-the-art for ImageNet training. Compared to \citet{akiba2017extremely}, our work saves around 40\% cost with similar hardware but much shorter training time. Compared to \citet{he2016deep}'s work which uses 8 GPUs, we achieve more than 248x speedup. Based on the same 1024 GPUs, our work is 1.61 times faster than \citet{akiba2017extremely}. Note that for ResNet-50 training, we adopt half-precision communication during the all-reduce gradients aggregation phase due to its reduced memory usage.  

For AlexNet, previously, \citet{you2017imagenet} could finish the ImageNet training with 32K mini-batch size in 11 minutes. As shown in Table~\ref{tb-alexnet}, we break this record and finish the AlexNet training in 4 minutes with 1024 Tesla P40 GPUs.

\subsection{Convergence Analysis}
\begin{figure}[!htbp]
    \centering
    \includegraphics[width=\linewidth]{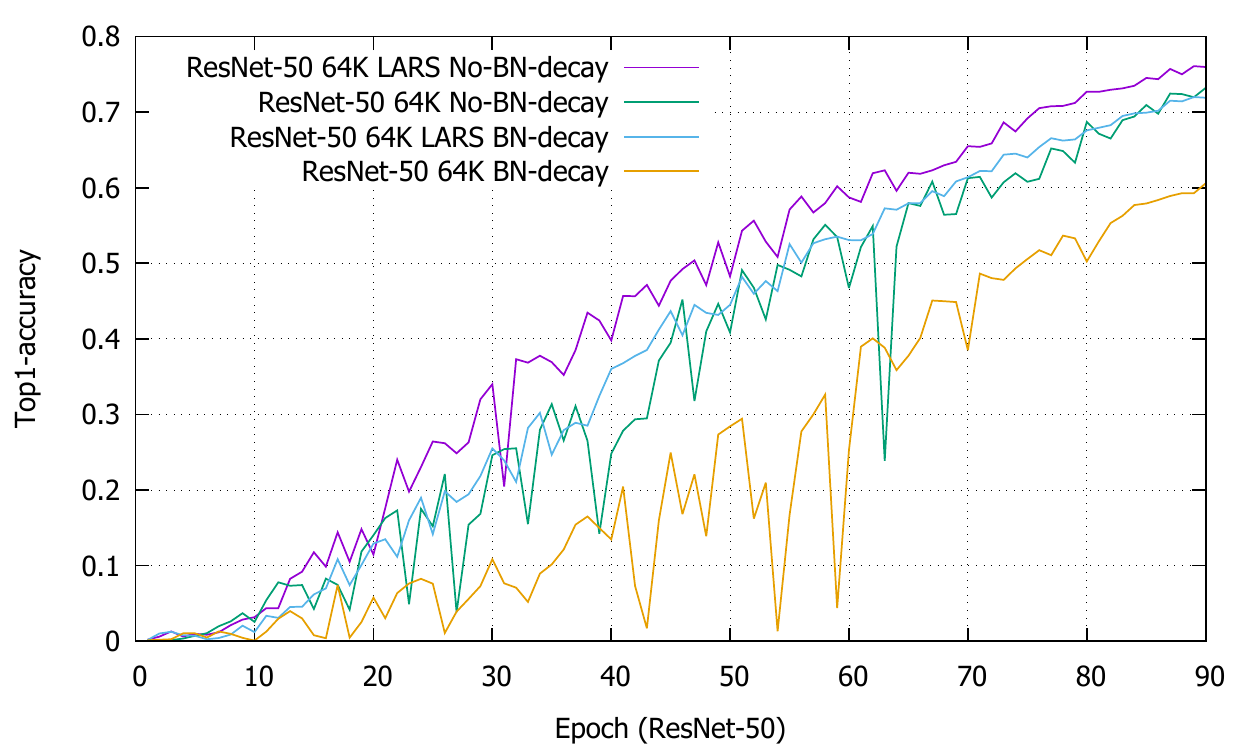}
    \caption{ImageNet Training with ResNet-50 Using 64K Mini-Batch Size}
    \label{fig:resnet-64k}
\end{figure}

\begin{figure}[!htbp]
    \centering
    \includegraphics[width=\linewidth]{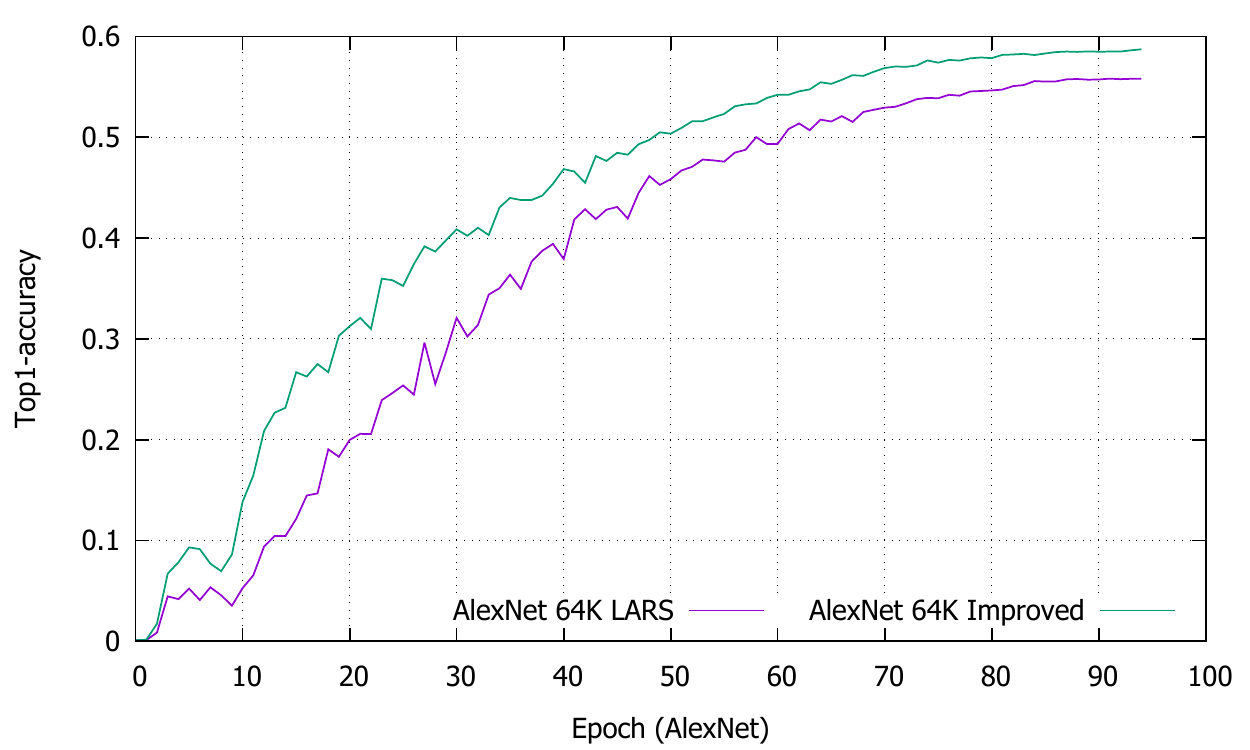}
    \caption{ImageNet Training with AlexNet Using 64K Mini-Batch Size}
    \label{fig:alexnet-64k}
\end{figure}

In this subsection, we show that with our optimizations, we can maintain the same convergence as previous works on ImageNet training with a larger mini-batch size. The overall training curve of top-1 accuracy for ResNet-50 and AlexNet are shown in Figure~\ref{fig:resnet-64k} and Figure~\ref{fig:alexnet-64k} separately.

% \begin{figure}
% 	\centering
% 	\begin{subfigure}{0.4\textwidth} % width of left subfigure
% 		\includegraphics[width=\textwidth]{resnet-acc}
% 		\caption{ResNet-50} % subcaption
% 		\label{fig:resnet-64k}
% 	\end{subfigure}
% 	\vspace{1em} % here you can insert horizontal or vertical space
% 	\begin{subfigure}{0.4\textwidth} % width of right subfigure
% 		\includegraphics[width=\textwidth]{alexnet-acc}
% 		\caption{AlexNet} % subcaption
%         \label{fig:alexnet-64k}
% 	\end{subfigure}
% 	\caption{ImageNet Training Using 64K Batch Size} % caption for whole figure
% \end{figure}

\begin{figure}[ht]
    \centering
    \includegraphics[width=0.5\textwidth]{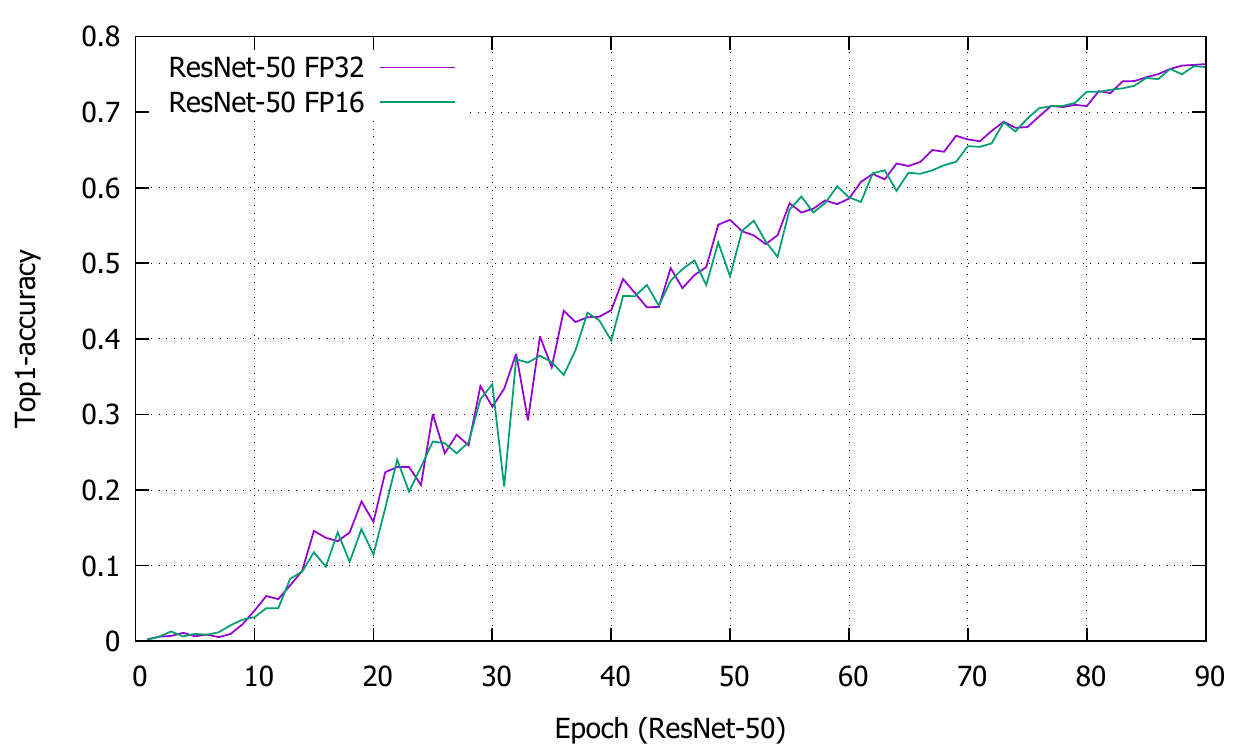}
    \caption{Compare the convergence of mixed-precision and single-precision training}
    \label{fig:mix-fp32}
\end{figure}

\textbf{Compare the convergence of mixed-precision training and single-precision training}.
As explained in Section~\ref{sec:Mixed-Precision}, we adopt a mixed-precision training strategy to avoid the precision loss in half-precision computation. A master copy of weights was updated in FP32 to avoid loss of accuracy, and all tensors in the forward and backward passes were in FP16. The updated gradients (weight gradients multiplied by the learning rate) become too small to be represented in FP16 as any value whose magnitude is smaller than $2^{-24}$ becomes zero in FP16. In fact, when the mini-batch size is less than 16K, a master copy of FP32 is enough to get the same top-1 test accuracy of baseline. With the mini-batch size of 16K, loss-scaling is required to maintain the same accuracy as the baseline, or else gradients vanishing will start to appear. When the mini-batch size increases to 32K, LARS technique was required for successful mixed precision training. To make LARS perform properly, we have to set its coefficient to a small number: 0.001. This will cause the local learning rate to become zeroes in FP16. Because of this, we need to assign an FP32 copy to LARS\@. Also, in order to avoid overfitting, the weight decay should be increased from 0.0001 to 0.0005 when the mini-batch size grows to 64K. 
To validate the effectiveness of our mixed-precision training strategy, we compare it with plain single-precision training. The experiment result in Figure~\ref{fig:mix-fp32} shows that our mixed-precision training strategy has similar top-1 accuracy for ResNet-50 at 90 epoch as single-precision training (76.3\% for single-precision training and 76.2\% for mixed-precision training).

\textbf{Effect of LARS.} We compare the test accuracy of ResNet-50 with and without applying LARS~\cite{you2017Scaling}. As shown in Table~\ref{tb:resnet-lars}, LARS could improve the top-1 accuracy from 60.6\% to 71.9\%. Also, Figure~\ref{fig:resnet-64k} shows that with LARS, the training curve is more smooth than the curve without LARS. However, even with both mixed-precision and LARS, we still cannot reach the baseline accuracy yet.

\begin{table}[htbp]
  \caption{Effect of LARS to ResNet-50 Training}
  \label{tb:resnet-lars}
  \centering
  \begin{tabular}{ccccc}
    \toprule
  Batch    & LARS & Top-1 Accuracy\\
    \midrule
  64K    & $\times$ & 60.6\%   \\
  64K      & \checkmark  & 71.9\% \\
    \bottomrule
  \end{tabular}
\end{table}

\textbf{Effect of model improvements.}
Eliminating weight decay on bias and batch normalization generates positive effects on convergence. Table~\ref{tb-resnet-factor} shows that eliminated weight decay on batch normalization(BN) for ResNet-50, combined with mixed-precision and LARS, could improve top-1 accuracy from 71.9\% to 76.2\%, which meets the baseline test accuracy. Note that for ResNet-50 training, we ignore the bias tensor for weight decay as its influence is negligible.

\begin{table}[htbp]
  \caption{Effect of improvements to ResNet-50 Training}
  \label{tb-resnet-factor}
  \centering
  \begin{tabular}{ccccc}
    \toprule
   Batch    & No Decay BN & Top1 \\
    \midrule
  64K      & $\times$ & 71.9\% \\
  64K    & \checkmark & 76.2\% \\
    \bottomrule
  \end{tabular}
\end{table}

% \begin{table}[htbp]
%   \caption{Effect of improvements to ResNet-50 Training}
%   \label{tb-resnet-factor}
%   \centering
%   \begin{tabular}{ccccc}
%     \toprule
%   Batch    & LARS & No Decay BN & Top1 \\
%     \midrule
%   64K    & $\times$ & $\times$ & 60.6\%   \\
%   64K    & $\times$ & \checkmark & 73.2\% \\
%   64K      & \checkmark & $\times$ & 71.9\% \\
%   64K    & \checkmark & \checkmark & 76.2\% \\
%     \bottomrule
%   \end{tabular}
% \end{table}

% \begin{table}[htbp]
%   \caption{Effect of improvements to AlexNet Training}
%   \label{tb-alexnet-factor}
%   \centering
%   \begin{tabular}{ccccc}
%     \toprule
%   Batch    & No Decay Bias & No Decay BN & Top1 \\
%     \midrule
%   64K    & $\times$ & $\times$ & 55.8\%   \\
%   64K    & $\times$ & \checkmark & 56.3\% \\
%   64K      & \checkmark & $\times$ & 56.4\% \\
%   64K    & \checkmark & \checkmark & 57.1\% \\
%   64K     & \checkmark & \checkmark & 58.8\% \\
%     \bottomrule
%   \end{tabular}
% \end{table}

\begin{table}[htbp]
  \caption{Effect of improvements to AlexNet Training}
  \label{tb-alexnet-factor}
  \centering
  \begin{tabular}{cccccc}
    \toprule
  Batch    & No Decay Bias & No Decay BN & pool5 BN & Top-1 Accuracy\\
    \midrule
  64K    & $\times$ & $\times$ & $\times$ & 55.8\%   \\
  64K    & $\times$ & \checkmark & $\times$ & 56.3\% \\
  64K      & \checkmark & $\times$ & $\times$ & 56.4\% \\
  64K    & \checkmark & \checkmark & $\times$ & 57.1\% \\
  64K     & \checkmark & \checkmark & \checkmark & 58.8\% \\
    \bottomrule
  \end{tabular}
\end{table}

For AlexNet, we test the effect of optimization strategies including not regularizing bias, not regularizing batch norm parameters and inserting batch normalization after Pool5 layer. As shown in Table~\ref{tb-alexnet-factor}, when applying all strategies, the top-1 accuracy of mini-batch size 64K reaches its peak value of 58.8\%, which meets the baseline accuracy. Figure~\ref{fig:alexnet-64k} also shows that after applying a series of optimization strategies, the convergence speed gets improved and the final test accuracy is higher than using the LARS algorithm only.

\subsection{Training Speed and Scalability}
In this subsection, we show the training speed and scalability of our distributed training system.

\textbf{Compare the speed of mixed-precision training and single-precision training}. As shown in Table~\ref{tb:compare-resnet-fp16}, using mixed-precision training can speedup single-node performance of ResNet-50 from 172 images/second to 218 images/second. This improvement comes from the FP16 computation speedup and the reduced communication parameter size.

\begin{table}[htbp]
  \caption{ResNet-50: Compare the speed of mixed-precision training and single-precision training}
  \label{tb:compare-resnet-fp16}
  \centering
  \begin{tabular}{ccc}
    \toprule
  Batch/GPU    & Data Type & Images/Sec \\
    \midrule
    64 & FP32 & 172 \\
    64 & mixed & 218 \\
    \bottomrule
  \end{tabular}
\end{table}

\begin{figure}[ht]
    \centering
    \includegraphics[width=\linewidth]{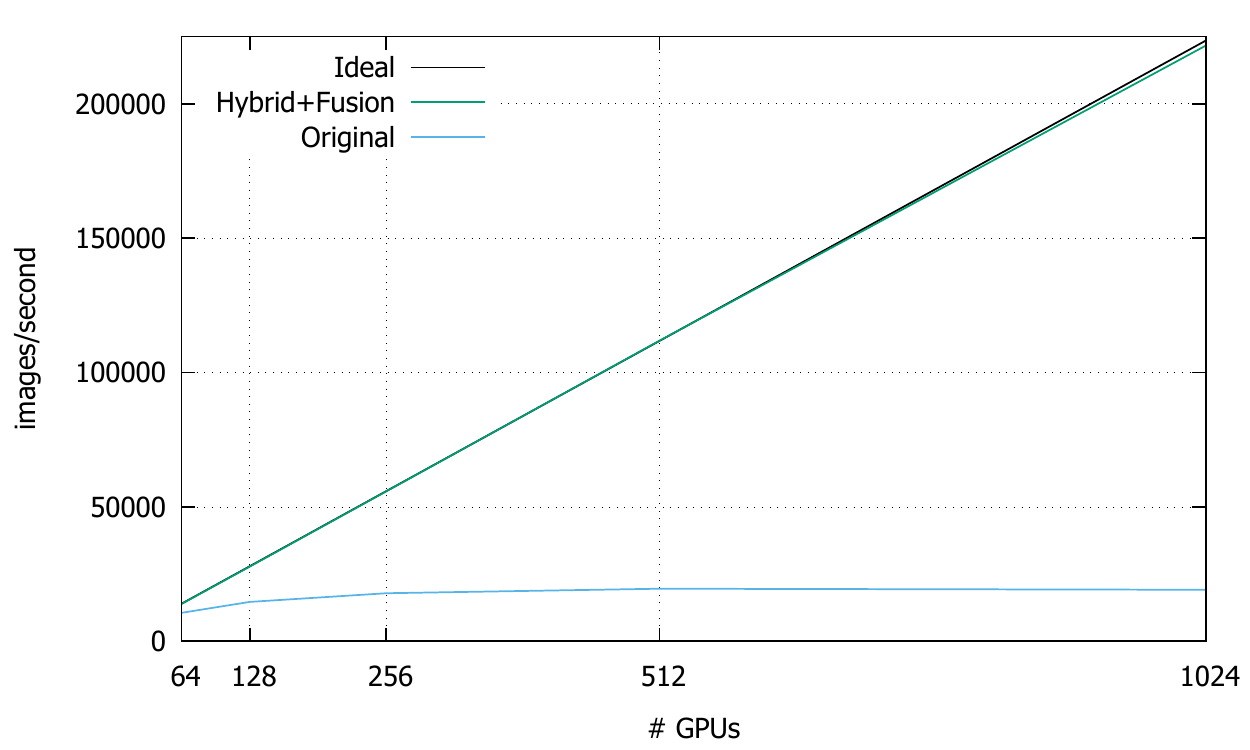}
    \caption{ResNet-50 training throughput with batch 64/GPU}
    \label{fig:resnet-scalability}
\end{figure}

\textbf{Scalability}. 
Figure~\ref{fig:resnet-scalability} shows that our customized all-reduce has high scaling efficiency. When per GPU mini-batch size is fixed to 64, the scaling efficiency of 1024 GPUs (8 GPUs $/product$ 128 nodes) compared to single-node (8 GPUs) could reach 99.2\%, which is close to the optimal scalability. When comparing the scaling efficiency before and after optimization, we can see the improvements is significant. For 1024 GPUs, we improved the scaling efficiency from 9.0\% to 99.2\%.

\begin{figure}[ht]
    \centering
    \includegraphics[width=\linewidth]{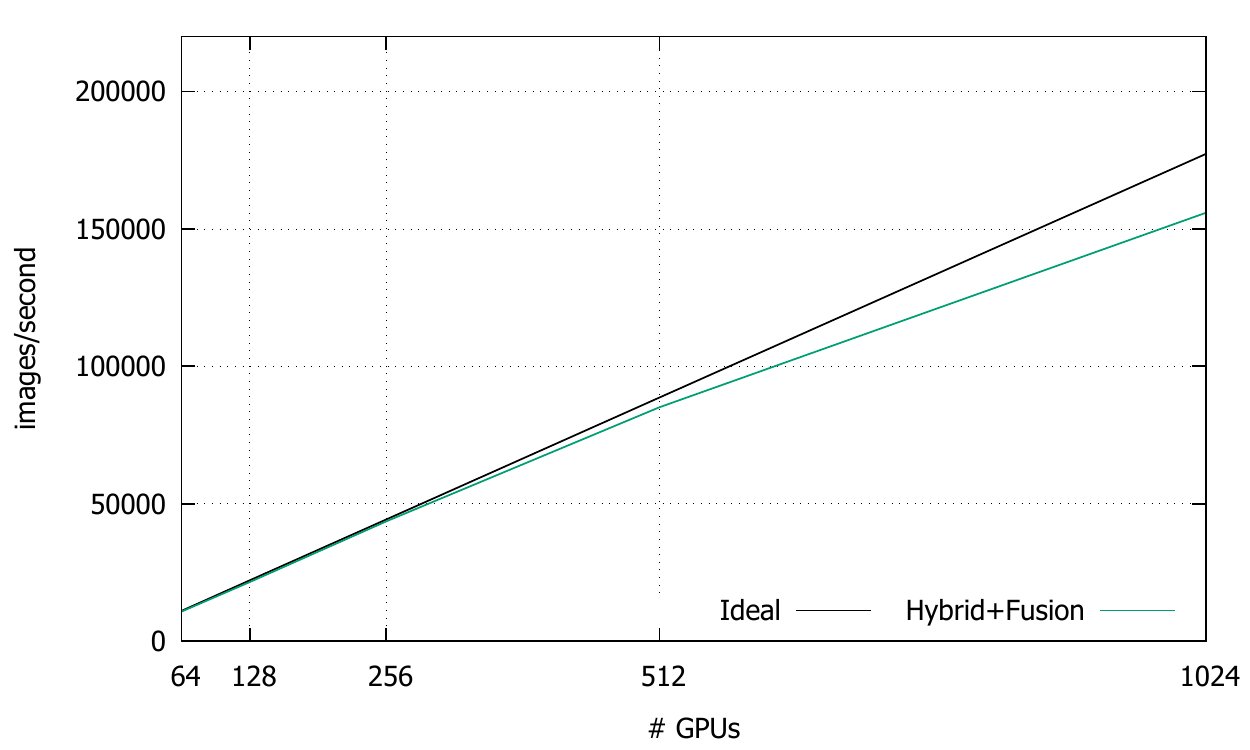}
    \caption{ResNet-50 training throughput with batch 32/GPU}
    \label{fig:resnet-scalability-32}
\end{figure}

When per GPU mini-batch size is fixed to 32, it is harder to scale out. Because the smaller mini-batch size often leads to faster computation, which causes the communication to become the bottleneck. As reported in~\cite{akiba2017extremely}, the scaling efficiency of 1024 GPUs with 32 batch/GPU is 80.0\%. As shown in Figure~\ref{fig:resnet-scalability-32}, our system can reach 87.9\% for the same batch settings as~\cite{akiba2017extremely}. Due to our efficient communication strategies, we have achieved higher scaling efficiency than the state-of-the-art with the same mini-batch size.

\begin{figure}[htbp]
    \centering
    \includegraphics[width=\linewidth]{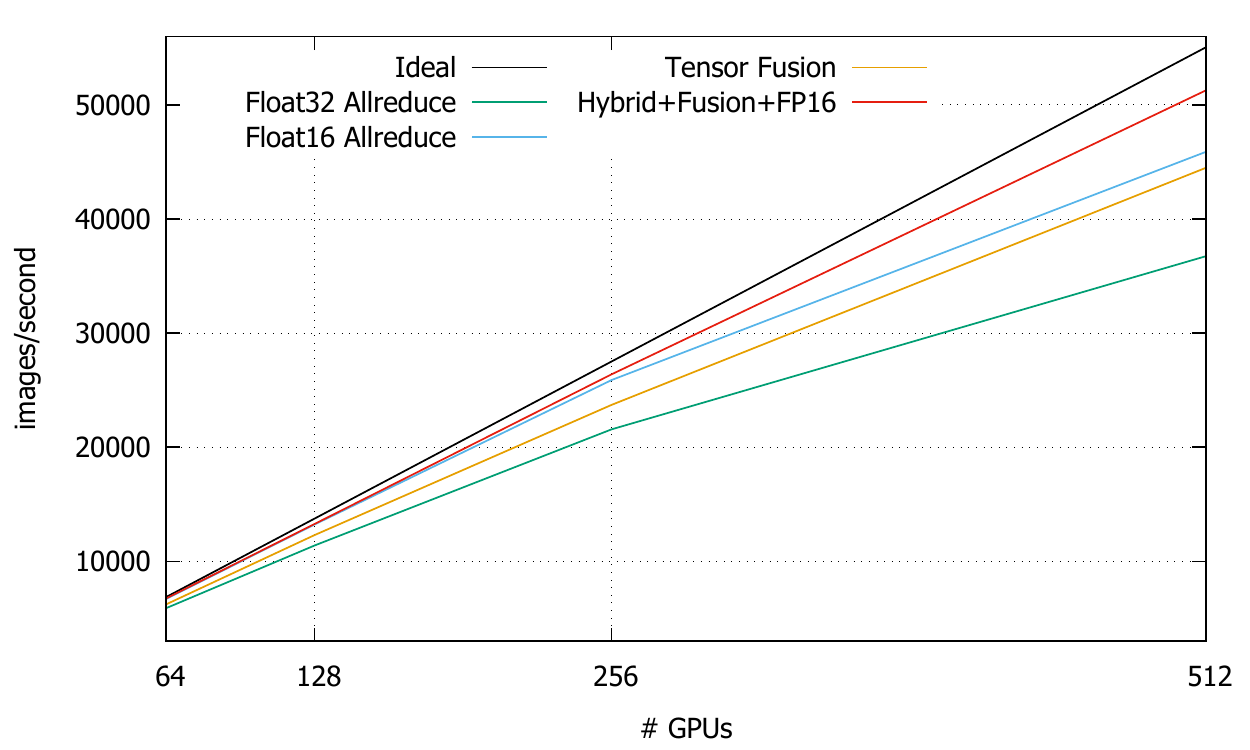}
    \caption{AlexNet training throughput with batch 128/GPU}
    \label{fig:alexnet-scalability}
\end{figure}

Figure~\ref{fig:alexnet-scalability} shows the scalability of AlexNet training with a mini-batch size of 128 per GPU\@. The baseline is FP32 all-reduce (with RDMA). When comparing the scaling efficiency between using 8 GPUs and 512 GPUs, introducing tensor fusion could achieve an improvement from 70\% to 81\%, and using FP16 all-reduce gives 82.7\% scalability. When combining FP16 and tensor fusion strategies with hybrid all-reduce, we get \textbf{91.4\%} scaling efficiency.

\section{Conclusion}
\label{sec:conc}
%To make full use of the computing capability of the fastest supercomputer for deep neural networks training, there are two challenges: 1) increasing the batch size while maintaining the test accuracy; and 2) making an extensible system with more GPUs while keeping the close-to-linear scalability. In this paper, we have built an efficient high-performance distributed system for training neural networks using distributed synchronized SGD\@. With optimizations on training strategy, model architecture, and communication algorithm, our system has achieved the \textbf{fastest} ImageNet training speed without losing accuracy. Our TensorFlow-based system can finish the \textbf{95-Epoch AlexNet training in 4 minutes with 58.7\% top-1 accuracy(batch size = 64K)}, and it can finish the \textbf{90-Epoch ResNet-50 training in 7 minutes with 76.18\% top-1 accuracy(batch size = 64K)}. 
% \slashwang{both versions of conclusion repeat what we have in abstract/introduction. Consider removing the conclusion part?}
Large-scale deep neural networks require a large amount of computation to converge to a good testing accuracy. Synchronized gradient descent methods with data parallelism are widely used to train the models in distributed environments. However, data communication between machines in the cluster easily becomes the bottleneck of the system throughput. Though using a large mini-batch size can improve the scalability of the system, it becomes more difficult to keep the good generalization ability of the models. In this study, we build a highly scalable deep learning training system to address the problem. We first use the mixed-precision techniques to improve the throughput of a single GPU without losing accuracy. Then we propose optimization approaches (e.g., eliminated weigh decay in batch normalization layers) to successfully train AlexNet and ResNet-50  using a mini-batch size of 64K without losing accuracy. To further increase the scalability of the system, we propose highly optimized all-reduce algorithms which achieve much better performance than the NCCL-based counterpart. As a result, on the training of the ImageNet dataset, we achieve 58.7\% top-1 test accuracy with AlexNet (95 epochs) in only 4 minutes using 1024 Tesla P40 GPUs, and achieve 75.8\% top-1 test accuracy with ResNet-50 (90 epochs) in only 6.6 minutes using 2048 Tesla P40 GPUs, which outperforms the existing systems. 

\bibliographystyle{ACM-Reference-Format}
\bibliography{imagenet-training}

\end{document}